\documentclass{article}

\RequirePackage{natbib}
\usepackage{graphicx} 
\usepackage[utf8]{inputenc} 
\usepackage[T1]{fontenc}    
\usepackage{hyperref}       
\usepackage{url}            
\usepackage{booktabs}       
\usepackage{amsfonts}       
\usepackage{nicefrac}       
\usepackage{microtype}      
\usepackage{listings}       
\usepackage{mathptmx}       

\usepackage[margin=0.5cm]{caption} 

\newcommand{\specialcellcenter}[2][c]{%
  \begin{tabular}[#1]{@{}c@{}}#2\end{tabular}}

\setcitestyle{authoryear,round,citesep={;},aysep={,},yysep={;}}

\renewcommand{\cite}[1]{\citep{#1}}

\usepackage{blindtext}
\usepackage{geometry}
\begin{document}

{\fontfamily{ptm}\selectfont
\title{Pretrained Generative Language Models as General \\ Learning Frameworks for Sequence-Based Tasks}}

\author{Ben Fauber\thanks{Correspondence to: Ben.Fauber@dell.com} \\
\normalsize{Dell Technologies}
}
\date{January 30, 2024}

\maketitle

\begin{abstract}
We propose that small pretrained foundational generative language models with millions of parameters can be utilized as a general learning framework for sequence-based tasks. Our proposal overcomes the computational resource, skill set, and timeline challenges associated with training neural networks and language models from scratch. Further, our approach focuses on creating small and highly specialized models that can accurately execute a challenging task of which the base model is incapable of performing. We demonstrate that 125M, 350M, and 1.3B parameter pretrained foundational language models can be instruction fine-tuned with 10,000-to-1,000,000 instruction examples to achieve near state-of-the-art results on challenging cheminformatics tasks. We also demonstrate the role of successive language model fine-tuning epochs on improved outcomes, as well as the importance of both data formatting and pretrained foundational language model selection for instruction fine-tuning success.
\end{abstract}

\section{Introduction}
Recent advances in generative large language models (LLMs) containing billions-to-trillions of model parameters have demonstrated a remarkable ability to create human-like text \cite{Achiam2023GPT4TR, Fedus2021SwitchTS}. Fine-tuning of LLMs has arisen as the preferred method to impart domain-specific information \cite{Chalkidis2020LEGALBERTT, Lee2019BioBERTAP}, ontology \cite{Baldazzi2023FinetuningLE}, commonsense reasoning capabilities \cite{AlKhamissi2023OPTRET, West2023NovaCOMETOC, Kimura2022TowardBA}, and/or style-transfer into these models \cite{Ding2023EnhancingCL, Touvron2023Llama2O}.

Our primary objective was to create highly specialized models by leveraging small pretrained foundational language models. In pursuit of this objective, we utilized the predefined architectures of these models as flexible learning frameworks, drawing parallels to traditional statistical learning models and algorithms. In this paper, we demonstrate that small pretrained foundational language models (\emph{i.e.}, generative models with millions of parameters) can be utilized as a general learning framework for sequence-based tasks.

Our goal was to determine if small pretrained foundational language models can be instruction fine-tuned \cite{Wei2021FinetunedLM} to perform a novel task of which the base language model is entirely naïve and incapable. The benefit of our proposed approach is that users can leverage recent advances in language models and instruction fine-tuning \cite{Liesenfeld2023OpeningUC}, along with a task-specific data set, and only a few minutes of language model fine-tuning to create a small specialist language model capable of performing an entirely new task. Small, specialized language models convey an additional benefit in that they require less computational infrastructure to fine-tune and inference than orders of magnitude larger generalist LLMs \cite{Kaplan2020ScalingLF, Hestness2017DeepLS}.

It is reasonable to counter our proposed approach with a full-scale neural network, and/or language model, training campaign for a specialized task \cite{Wu2023BloombergGPTAL, Beltagy2019SciBERTAP}. \textbf{Yet, our proposed small pretrained foundational language model fine-tuning paradigm conveys three key benefits over a full-scale model training campaign:}
\begin{enumerate}
\item Fewer computational resources required; 
\item Accessible to most data scientists and software engineers; and
\item Much faster, requires less time, thus accelerates the "\emph{design-make-test}" cycle and time-to-value.
\end{enumerate}

\subsection{Novel Specialized Task}

It has been noted that LLMs perform well in the generation of human-like text and many generalist applications \cite{Achiam2023GPT4TR, Anil2023PaLM2T}, yet they have continuously demonstrated deficiencies in scientific topics, especially mathematics and chemistry \cite{Lewkowycz2022SolvingQR, Cobbe2021TrainingVT, Hendrycks2021MeasuringMP}. As such, we chose a chemistry-based objective for our specialized small language model fine-tuning task. 

\begin{figure}[h!]
\begin{center}
\includegraphics[width=100mm]{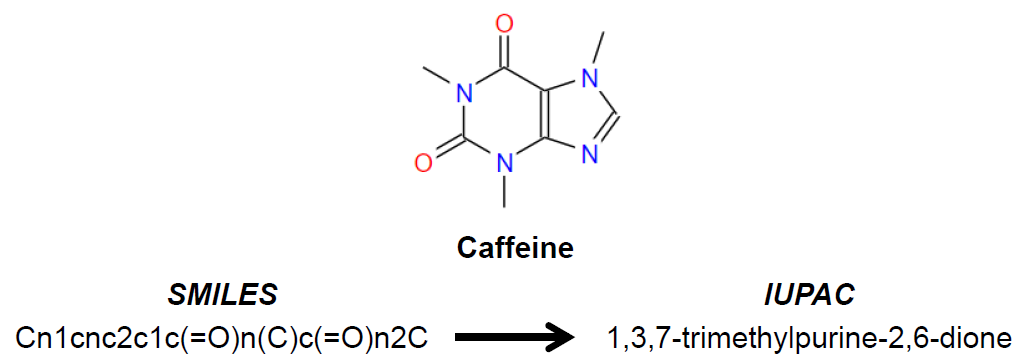}
\caption{Illustration of our proposed task: convert a molecular SMILES string into the corresponding IUPAC chemical name. Skeletal structure of caffeine with the SMILES string representation and corresponding IUPAC chemical name are shown as an example.}
\label{overview}
\end{center}
\end{figure}

Specifically, we chose to evaluate the conversion of a SMILES (Simplified Molecular-Input Line-Entry System) string \cite{Swanson2004TheEO, Weininger1988SMILESAC}, which is a cheminformatics representation of a molecular structure, into the molecule's corresponding IUPAC (International Union of Pure and Applied Chemistry) chemical name \cite{Favre2014NomenclatureOO, Fennell1994HistoryOI} (Figure 1). This is a challenging task for humans and only achievable on the simplest of molecules, with IUPAC naming of larger molecules and programmatic applications performed by rules-based processes (\emph{i.e.}, software from Daylight Chemical Information Systems, OpenEye Scientific Software, and/or MolSoft LLC). In our studies, we demonstrated that pretrained foundational language models cannot perform this task with any measurable proficiency (\emph{see} Appendix for details).

\section{Related Work}

Encoder-decoder neural networks have demonstrated reasonable performance in the conversion of SMILES strings into the corresponding IUPAC chemical names after training on more than 30 million examples for multiple days \cite{Rajan2020STOUTST}. Despite this development, the rules-based naming process remains state-of-the-art for this task. 

Further, a neural network was trained for the inverse task of converting IUPAC chemical names into the corresponding SMILES strings \cite{Krasnov2021TransformerbasedAN}. Again, this process required training on over 30 million examples for multiple weeks to achieve similar performance to a rules-based approach. Despite this development, a rules-based approach also remains state-of-the-art for this inverse task \cite{Lowe2011ChemicalNT}.

\subsection{Our Contribution}

We emphasize that our objective was not to create a superior model to either the rules-based nor encoder-decoder methodologies. Rather our goal was to determine:
\begin{enumerate}
    \item Can we achieve reasonable performance of this challenging task in comparison to the reported methods? \item How much domain-specific fine-tuning data is required?
    \item Which language model instruction fine-tuning conditions are preferred in transforming an ineffective pretrained foundational language model into a fine-tuned model capable of this task? 
\end{enumerate}

In our work, we used small pretrained foundational language models (\emph{i.e.}, generative models with millions of parameters) as starting points, along with domain-specific data for our task, allowed instruction fine-tuning of the foundational model for a few epochs (requiring anywhere from 2-to-90 minutes based on the size of the instruction data set and foundational model). We evaluated the performance of our instruction fine-tuned language models against ground truth data (\emph{i.e.}, hold-out "test" data set) in a rigorous and reproducible evaluation framework.

\section{Methods}

\subsection{Data Set}

Our parent data set was composed of approximately 30 million instances of molecular SMILES strings and their corresponding IUPAC names from the free and publicly accessible United States National Institutes of Health (NIH) PubChem database \cite{Kim2015PubChemSA}. We curated the PubChem data to create a parent dataset of organic molecules where all instances were unique and there were no duplicates of either SMILES strings or IUPAC chemical names. Details of the parent data set sourcing and curation are further described in the Appendix section.

Following best practices in traditional machine learning, we randomly divided the parent data set into fine-tuning data (80\% of the parent data) and test data (20\% of the parent data) by sampling without replacement. The fine-tuning data set was unique and distinct from the test data set.

We varied the number of available fine-tuning data instances from 100 to 10,000,000 examples of SMILES strings and their corresponding IUPAC chemical names, by random selection without replacement from the parent pool of fine-tuning data instances for each instance cohort. The fine-tuning data instances were used for language model instruction fine-tuning. The language models were never exposed to the test data (\emph{i.e.}, hold-out data) during the fine-tuning process to avoid train/test data contamination. 

All data was formatted into an instruction-based format where the “instruction” was the input, and the “output” was the desired outcome. For example the instruction was, “Translate the following SMILES string into an IUPAC name: CN=C(C1=C(N=C(N1C)C2CCCCC2)C3=CC4=C(N3)C=CNC4=O)N,” and the corresponding output was, “2-cyclohexyl-N',3-dimethyl-5-(4-oxo-1,5-dihydropyrrolo[3,2-c]pyridin-2-yl)imidazole-4-carboximidamide”. The instruction formatting was consistent throughout the fine-tuning and testing datasets.

\subsection{Selection of Pretrained Foundational Language Models}

Large language models have shown remarkable generalist capabilities in creating human-like prose, knowledge retention, as well as commonsense reasoning abilities \cite{Xi2023TheRA}. These high-performing generalist generative language models often possess >100 billion model parameters in addition to very large training data sets. Examples of these generalist LLMs include GPT-4 \cite{Achiam2023GPT4TR} and PaLM-2 \cite{Anil2023PaLM2T}. 

Conversely, we chose to utilize orders of magnitude smaller generative language models, such as those with approximately 100 million parameters. We selected the OPT (open pretrained transformer) family of pretrained foundational generative language models as the starting point for our studies \cite{Zhang2022OPTOP}. The OPT family of language models also conveyed the benefit of smaller and larger model sizes being available within the same model family and architecture. This allowed us to systematically evaluate the impact of changes in model size (\emph{i.e.}, parameter count) and performance against our specialized task.

In our work, we define model fine-tuning as initialization of a pretrained foundational language model followed by updates to the model weights and biases. In our fine-tuning setting, all language model parameters can undergo gradient updates --- there are no frozen layers nor adapters.

The prompt for the language models were consistent throughout our evaluation and across all models. The language model prompt was general and agnostic to the data set instructions. The prompt used for our evaluation was:
“Below is an instruction that describes a task. Write a response that appropriately completes the request. \#\#\# Instruction: \{instruction\} \#\#\# Response:”.

\subsection{Evaluation of our Method}

Evaluating the performance of fine-tuned language models on converting a SMILES string into the corresponding IUPAC chemical name was chosen due to its presumed limited exemplification in language model training corpuses. We correctly assumed that this task could not be adequately executed by pretrained foundational language models (\emph{see} Appendix). Additionally, this task offered an opportunity to concretely evaluate the instruction fine-tuned language model outputs against ground truth data with natural language processing (NLP) metrics.

IUPAC chemical names are not singularly definitive and there are likely multiple reasonable IUPAC names that could correspond to the SMILES string of a complex molecule \cite{Favre2014NomenclatureOO}. It is also reasonable to assume that multiple correct variants of an IUPAC chemical name exist within the chemical literature and are acceptable to chemists. 

Conversely, we set a high bar for our studies by viewing the ground truth IUPAC chemical name in our parent data set as the only acceptable language model output. This requirement is also in part due to the domain-specific fine-tuning data set that we created. Our parent data set contained only one IUPAC chemical name per SMILES string example. Additionally, this strict comparison framed the more general questions of: 1) How well does an instruction fine-tuned language model learn the required output when provided the task-specific instruction fine-tuning data? and 2) What are the domain-specific data and model fine-tuning requirements to achieve reasonable results?

We evaluated the performance of the pretrained foundational language models, as well as the instruction fine-tuned language models, against the ground truth test data sets using multiple NLP metrics. First, we evaluated the percentage of the language model outputs that exactly matched the ground truth. 

Second, we selected normalized Levenshtein edit distance as a reasonable NLP metric to compare the output of the language model against the ground truth data. Edit distance considers the number of deletions, substitutions, and transpositions required to transform one string into another \cite{Levenshtein1965BinaryCC}. Normalized edit distance is the edit distance between two strings, divided by the length of the longest string, thus bounded within $[0,1]$, where 1 is a perfect match \cite{Marzal1993ComputationON}. Normalized edit distance also allows for comparison of multiple results across a dataset when there is variation in string length.

Third, we employed the same prescriptive chunking and BLEU (BiLingual Evaluation Understudy) scoring $[0,1]$, where 1 is a perfect match \cite{Papineni2002BleuAM}, described in the encoder-decoder neural network method \cite{Rajan2020STOUTST}. The chunked model output was compared against the chunked ground truth via BLEU scoring, and this method also permitted comparison of our method versus the prior encoder-decoder methodology. Again, we recognize that the rules-based approach is state-of-the-art for this task, and our objective was not to create a superior model. Rather, our goal was to determine if the performance of our models were in line with the performance of the published encoder-decoder method via BLEU score comparisons.

Examples of SMILES strings and their corresponding IUPAC chemical names can be found in the Appendix section. Additionally, there are examples of normalized edit distance and BLEU scores variation with subtle changes in the instruction fine-tuned language model outputs relative to the ground truth data.

\subsection{Consistency in Evaluation Outcomes}

The pretrained foundational language models and the fine-tuned language models were evaluated against the same test data set. The same batch of test data was used for the evaluation of all the models for consistency and comparison of outcomes. The same language model fine-tuning and generation configurations were utilized throughout our studies, and only single-parameter changes were permitted, as annotated in the tables, when comparing methods. 

A single pretrained foundational language model and a single fine-tuned language model were evaluated against a single test data set, in triplicate, to determine text generation variability and its impact on downstream evaluation metric variability. There was no detectable variability in the evaluation metrics. 

Further, a single pretrained foundational language model was fine-tuned, in triplicate, using the same domain-specific data set to determine the variability in model fine-tuning, and the downstream influence on text generation and metric evaluation variability. Again, there was no detectable variability in the evaluation metrics. 

Finally, a single pretrained foundational language model was fine-tuned using the same domain-specific data set, using three discrete random seed values for the language model fine-tuning process. Our goal was to determine the variability in model fine-tuning with different random seeds, and the downstream impact on text generation and metric evaluation variability. In this setting there was no detectable variability in the normalized edit distance metrics and $\le$ 2\% variability in the percent exact match and mean BLEU score metrics. The details of these variability studies are further discussed in the Appendix section.

\section{Results}

\subsection{Baseline Performance}

Language models can be generalists, or specialists, and it is valuable to understand what is required to create a specialist language model. It is important to determine how much fine-tuning data, and which fine-tuning paradigm, are reasonable starting points to create a specialist language model. 

We recognize that fine-tuning language models over multiple epochs may obliterate some portion of information that resides within the pretrained foundational language model. This potential change did not concern us as our objective was to create specialized language models from pretrained foundational language models, with the objective of effectively executing a highly specialized task that the original pretrained foundation models were incapable of performing. 

We set the high bar of accurately converting SMILES strings into IUPAC chemical names and achieving exact matches to the ground truth in the test data set. Pretrained foundational language models are unable to perform this task with any reasonable level of proficiency (Table 1). Additionally, the pretrained foundational language models are incapable of achieve a measurable BLEU score on this task.

\begin{table*}[h!]
\begin{center}
\begin{small}
\begin{tabular}{lcccc}
\toprule
\specialcellcenter{Pretrained Foundational \\ Language Model} & \specialcellcenter{Language Model \\ Parameter Count} & \specialcellcenter{\% Exact \\ Matches} & \specialcellcenter{Mean Normalized \\ Edit Distance} & \specialcellcenter{Mean BLEU \\ Score} \\
\midrule
facebook/opt-125m & 125M & 0\% & 0.09 & 0 \\
facebook/opt-350m & 350M & 0\% & 0.07 & 0 \\
facebook/opt-1.3b & 1.3B & 0\% & 0.08 & 0 \\
facebook/opt-6.7b & 6.7B & 0\% & 0.03 & 0 \\
facebook/opt-13b & 13B & 0\% & 0.02 & 0 \\
facebook/opt-30b & 30B & 0\% & 0.01 &  0 \\
microsoft/phi-2 & 2B & 0\% & 0.07 & 0 \\
meta-llama/Llama-2-7b-hf & 7B & 0\% & 0.00 & 0 \\
meta-llama/Llama-2-13b-hf & 13B & 0\% & 0.04 & 0 \\
meta-llama/Llama-2-70b-hf & 70B & 0\% & 0.24 & 0 \\
mistralai/Mistral-7B-Instruct-v0.2 & 7B & 0\% & 0.11 & 0 \\
\bottomrule
\end{tabular}
\end{small}
\caption{Baseline performance of pretrained foundational language models in the conversion of 1,000 test instances of SMILES strings into IUPAC chemical names, and comparison of the model output to the ground truth. The language models are described by their \texttt{HuggingFace.co} repo names (accessed 30Dec2023).}
\end{center}
\end{table*}

\subsection{Influence of Data Set Size}

Our objective was to expose small pretrained foundational language models to increasing orders of magnitude of domain-specific instruction fine-tuning data and assess their performance against test data with well-defined metrics. We utilized three different metrics for our assessment: 1) percentage of exact matches against the ground truth, 2) normalized edit distance against the ground truth; and 3) BLEU score of the chunked IUPAC name against the chunked ground truth.

\begin{figure}[h!]
\begin{center}
\includegraphics[width=100mm]{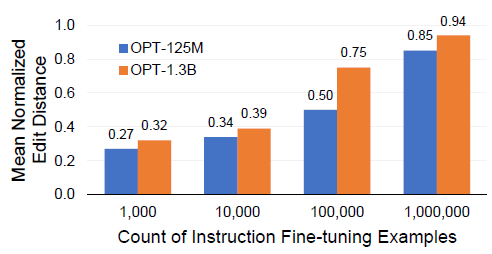}
\caption{Influence of increasing instruction fine-tuning examples. Mean normalized edit distance versus count of the instruction fine-tuning examples for the OPT-125M (blue) and OPT-1.3B (orange) language models, instruction fine-tuned with learning rate = 2e-5, batch size = 4, and epochs = 3. Performance of the instruction fine-tuned language models were assessed with 1,000 test instances of SMILES strings, and the model’s ability to accurately generate the corresponding IUPAC chemical names relative to the ground truth.}
\label{increasing_examples}
\end{center}
\end{figure}

We demonstrated that increasing the amount of domain-specific instruction fine-tuning data, as well as increasing the pretrained foundational language model parameter count (\emph{i.e.}, model size), improved the fine-tuned language model performance on our task (Figure 2). Further, the OPT-1.3B foundational language model instruction fine-tuned on 1,000,000 examples achieved a mean BLEU score of 0.92, meeting or exceeding the prior art for this task and metric \cite{Rajan2020STOUTST}.

It has been well documented that LLMs can perform more proficiently than smaller language models on challenging tasks \cite{Brown2020LanguageMA}. It has been postulated this increase in performance is due to the capability of larger language models to learn infrequent features due to their larger neural network and higher parameter count \cite{Wei2022EmergentAO, Hoffmann2022TrainingCL}.

\subsection{Increasing Fine-Tuning Epochs}

We found that increasing the number of instruction fine-tuning epochs associated with the OPT-125M language model fine-tuning process (Figure 3), substantially improved the ability for the instruction fine-tuned language model to perform a task for which the base model was incapable of achieving any reasonable performance. This result also demonstrates the ability to improve the instruction fine-tuned model performance for a task with a constrained amount of instruction fine-tuning data. Yet, the successive improvements in results from increasing epochs do appear to plateau after 20-30 fine-tuning epochs.

\begin{figure}[h!]
\begin{center}
\includegraphics[width=100mm]{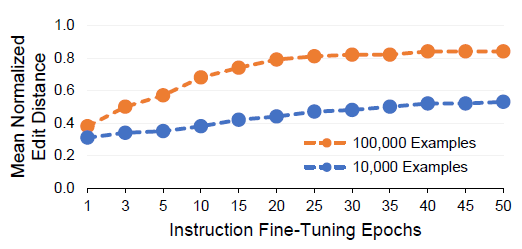}
\caption{Influence of increasing instruction fine-tuning epochs. Mean normalized edit distance versus instruction fine-tuning epochs for 10,000 (blue) and 100,000 (orange) instruction fine-tuning examples and the OPT-125M pretrained foundational language model, instruction fine-tuned with learning rate = 2e-5 and batch size = 4. Performance of the instruction fine-tuned language models were assessed with 1,000 test instances of SMILES strings, and the model’s ability to accurately generate the corresponding IUPAC chemical names relative to the ground truth.}
\label{increasing_epochs_ed}
\end{center}
\end{figure}

Likewise, we observed successive improvements in the percentage of exact matches achieved by increasing the number of instruction fine-tuning epochs associated with the OPT-125M language model fine-tuning process (Figure 4). Again, our process substantially improved the ability of the fine-tuned language model to perform a task for which the base model was incapable of achieving any detectable performance. In this instance, at least 100,000 instruction fine-tuning examples were required to achieve reasonable performance against the highest-bar metric of percentage exact matches. The successive improvements via increased fine-tuning epochs do appear to plateau after approximately 40 instruction fine-tuning epochs.

\begin{figure}[h!]
\begin{center}
\includegraphics[width=100mm]{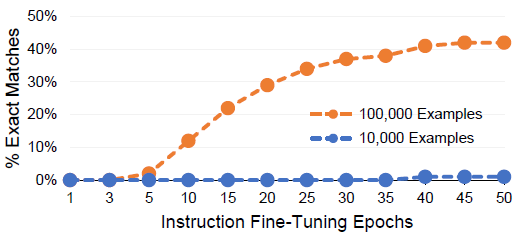}
\caption{Influence of increasing instruction fine-tuning epochs on \% exact matches. Percentage of exact matches versus instruction fine-tuning epochs for 10,000 (blue) and 100,000 (orange) instruction fine-tuning examples and the OPT-125M pretrained foundational language model, instruction fine-tuned with learning rate = 2e-5 and batch size = 4. Performance of the instruction fine-tuned language models were assessed with 1,000 test instances of SMILES strings, and the model’s ability to accurately generate the corresponding IUPAC chemical names relative to the ground truth.}
\label{increasing_epochs_matches}
\end{center}
\end{figure}

\subsection{Influence of Language Model Size}

We also demonstrated that systematically increasing the model parameter count within the OPT family of pretrained foundational language models improved outcomes. The fine-tuned language models with higher parameter counts performed better than the smaller parameter count models when exposed to the same instruction fine-tuning data set and fine-tuning process, as assessed by the percentage of exact matches (Figure 5). Notably, instruction fine-tuning the OPT-1.3B model with 1,000,000 examples resulted in 71\% exact matches.

\begin{figure}[h!]
\begin{center}
\includegraphics[width=100mm]{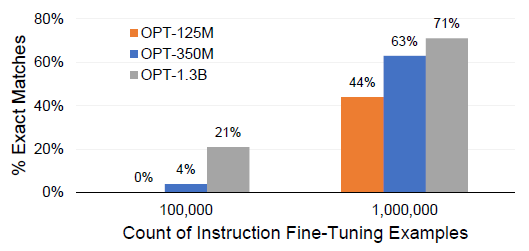}
\caption{Influence of increasing instruction fine-tuned language model parameter count on \% exact matches. Percentage of exact matches versus 100,000 and 1,000,000 instruction fine-tuning examples for the OPT-125M (orange), OPT-350M (blue), and OPT-1.3B (grey) pretrained foundational language models, instruction fine-tuned with learning rate = 2e-5, batch size = 4, and epochs = 3. Performance of the instruction fine-tuned language models were assessed with 1,000 test instances of SMILES strings, and the model’s ability to accurately generate the corresponding IUPAC chemical names relative to the ground truth.}
\label{increasing_model_size}
\end{center}
\end{figure}

Additionally, our systematic model fine-tuning and evaluation process revealed substantial improvements in the BLEU score of the chunked IUPAC chemical name model outputs against the chunked ground truth (Figure 6). The performance of our OPT-1.3B pretrained foundational model instruction fine-tuned with 1,000,000 examples was comparable, or superior to, the performance of other encoder-decoder networks for this task \cite{Rajan2020STOUTST}. Yet, our approach required less than 1 hour or fine-tuning, and a fraction of the data that was utilized for the construction of the encoder-decoder network. Our results further highlight the simplicity and utility of pretrained generative language models as general frameworks for learning sequence-based tasks.

\begin{figure}[h!]
\begin{center}
\includegraphics[width=100mm]{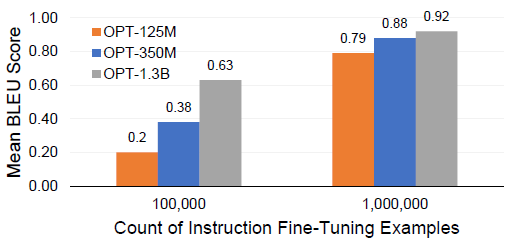}
\caption{Influence of increasing instruction language model parameter count on mean BLEU score. Mean BLEU score versus 100,000 and 1,000,000 instruction fine-tuning examples for the OPT-125M (orange), OPT-350M (blue), and OPT-1.3B (grey) pretrained foundational language models, instruction fine-tuned with learning rate = 2e-5, batch size = 4, and epochs = 3. Performance of the instruction fine-tuned language models were assessed with 1,000 test instances of SMILES strings, and the model’s ability to accurately generate the corresponding IUPAC chemical names relative to the ground truth.}
\label{increasing_model_size_bleu}
\end{center}
\end{figure}

\subsection{Smaller Generative Language Models}

We were inspired by the \texttt{TinyStories} results where very small language models (\emph{e.g.}, 1M-to-33M parameters) were trained on simple text, like language that a three- or four-year-old would understand \cite{Eldan2023TinyStoriesHS}. This resulted in small generative language models that could create simplistic yet readable text. Encouraged by the performance of these small generative language models, we further explored them in the context of our specialized task.

We fine-tuned the TinyStories family of small pretrained language models on different cohorts of instruction fine-tuning data sets and assessed their abilities to convert SMILES strings into the corresponding IUPAC chemical names. The TinyStories family of models also conveyed the benefit of simultaneously exploring the impact of model architecture on model performance, as the family contains several examples of similar/same language model parameter counts while varying the number of layers (\emph{i.e.}, depth) of the models.

\begin{figure}[h!]
\begin{center}
\includegraphics[width=120mm]{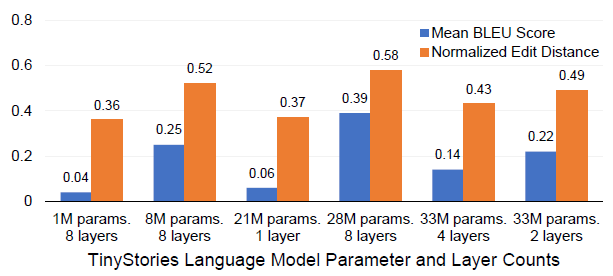}
\caption{Illustration of mean BLEU scores (blue) and mean normalized edit distance scores (orange) versus the TinyStories small pretrained foundational language model family by model parameter count (params.) and number of layers within the models. All foundational models were instruction fine-tuned on 1,000,000 examples with learning rate = 2e-5, batch size = 4, and epochs = 3. Performance of the instruction fine-tuned language models were assessed with 1,000 test instances of SMILES strings, and the model’s ability to accurately generate the corresponding IUPAC chemical names relative to the ground truth.}
\label{tinystories}
\end{center}
\end{figure}

Our results in the instruction fine-tuning of the TinyStories family of small pretrained foundational language models demonstrated that very small pretrained generative language models (\emph{e.g.}, 1M parameters) can learn a novel task with a large (\emph{e.g.}, 1M examples) instruction fine-tuning data set (Figure 7). 

The performance of the instruction fine-tuned small language models followed a similar trend when assessed by either mean BLEU score or mean normalized edit distance (Figure 7). Notably, the performance of the 28M parameter 8-layer instruction fine-tuned language model was optimal, with this instance also achieving 6\% exact matches on the test data – the best result of all the instruction fine-tuned TinyStories language models in our study. 

The 28M parameter fine-tuned model outperformed the 33M parameter instances with 2 and 4 layers, respectively. Thus, demonstrating that the model parameter count alone does not determine the ability of a small pretrained foundational language model to absorb instruction fine-tuning data. Rather, language model parameter count, and the model architecture are both considerations, where the deeper (\emph{i.e.}, 8 layer) models appeared to perform better than shallow models (\emph{e.g.}, 1, 2, or 4 layers).

The results we achieved by fine-tuning the TinyStories family of small foundational language models on 1,000,000 instruction examples were superior to those achieved with only 100,000 instruction examples (data not shown). We note that the TinyStories family of small pretrained foundational language models, without instruction fine-tuning, were incapable of performing our task at any measurable level of proficiency (\emph{see} Appendix).

\subsection{Other Language Models}

The promising results from instruction fine-tuning foundational language models of 1M-to-1.3B parameters led us to explore other models of similar size to assess the generality of our approach across various pretrained foundational language models. None of the pretrained foundational language models we explored could perform our task at any measurable level of proficiency without additional instruction fine-tuning on our domain-specific data sets (Table 2).

Most pretrained foundational language models of $\sim$125M parameters follow the architecture established by GPT-2 (Generative Pre-trained Transformer) \cite{Radford2019LanguageMA}. GPT-neo-125M \cite{Black2021GPTNeoLS}, GPT-2-small, and OPT-125M all have $\sim$125M model parameters along with 12 layers, 12 attention heads, and 768-dimension embedding architecture. 

Despite the similarities in architecture, the performance of the $\sim$125M parameter language models instruction fine-tuned on our 100,000 and 1,000,000 example instruction fine-tuning data sets were somewhat different (Table 2). Notably, GPT-2 outperformed GPT-neo, but OPT-125M was still the best fine-tuned language model of this size across all evaluation metrics (\emph{see} Figures 2, 5, and 6).

\begin{table*}[h!]
\begin{center}
\begin{small}
\begin{tabular}{lccccc}
\toprule
\specialcellcenter{Pretrained \\ Foundational \\ Language Model} & \specialcellcenter{Language \\ Model \\ Parameters} & 
\specialcellcenter{Instruction \\ Fine-Tuning \\ Dataset Size} & \specialcellcenter{\% Exact \\ Matches} & \specialcellcenter{Mean \\ Normalized \\ Edit Distance} & \specialcellcenter{Mean \\ BLEU \\ Score} \\
\midrule
GPT-neo-125m & 125M & Base Model & 0\% & 0.08 & 0 \\
GPT-neo-125m & 125M & 100,000 & 0\%	& 0.39 & 0.07 \\
GPT-neo-125m & 125M & 1,000,000	& 2\% & 0.53 & 0.29 \\
GPT-2 (small) & 124M & Base Model & 0\% & 0.06 & 0 \\
GPT-2 (small) & 124M & 100,000 & 0\%	& 0.39 & 0.06 \\
GPT-2 (small) & 124M & 1,000,000 & 7\% & 0.63 & 0.44 \\
Bloom-1b1 & 1.1B & Base Model & 0\% & 0.05 & 0 \\
Bloom-1b1 & 1.1B & 100,000 & 28\% & 0.77 & 0.67 \\
Bloom-1b1 & 1.1B & 1,000,000 & 68\%	& 0.93 & 0.90 \\
SantaCoder & 1.1B & Base Model & 0\% & 0.04 & 0 \\
SantaCoder & 1.1B & 100,000 & 24\% & 0.77 & 0.65 \\
SantaCoder & 1.1B & 1,000,000 & 75\% & 0.95 & 0.94 \\
\bottomrule
\end{tabular}
\end{small}
\caption{Baseline and instruction fine-tuning performance of other language models. Pretrained foundational language models were instruction fine-tuned with 100,000 and 1,000,000 examples where learning rate = 2e-5, batch size = 4, and epochs = 3. Performance of baseline models and the instruction fine-tuned language models were assessed with 1,000 test instances of SMILES strings, and the model’s ability to accurately generate the corresponding IUPAC chemical names relative to the ground truth.}
\end{center}
\end{table*}

The Bloom-1b1 \cite{Scao2022BLOOMA1} and SantaCoder \cite{Allal2023SantaCoderDR} 1.1B parameter pretrained foundational language models were also instruction fine-tuned on our 100,000 and 1,000,000 example data sets. These two language models and OPT-1.3B shared a similar model architecture of 24 layers, 16 or 32 attention heads, and a 1,536- or 2,048-dimension embedding space. 

The performance of the $\sim$1B parameter language models when instruction fine-tuned on our 100,000 and 1,000,000 example data sets differed despite their similar architectures (Table 2). Notably, the Bloom-1b1 and SantaCoder instruction fine-tuned models outperformed the instruction fine-tuned OPT-1.3B models (\emph{see} Figures 2, 5, and 6). Further, the Bloom-1b1 and SantaCoder foundational language models instruction fine-tuned on 1,000,000 examples achieved mean BLEU scores of 0.90 and 0.94, respectively, meeting or exceeding the prior art for this task and metric \cite{Rajan2020STOUTST}.

The observed differences in instruction fine-tuned model performance, despite similar model parameter counts and architectures might be attributable to subtle differences in attention strategies associated with each model (\emph{e.g.}, global, global-local, self, and multi-query). The different outcomes we observed may also be attributable to the language model instruction fine-tuning process paired with the information inherent in the pretrained foundational models. Thus, the information within the pretrained foundational model might be important in achieving high-quality language model instruction fine-tuning results, even for highly specific tasks for which the base model is naïve.

\subsection{Fine-Tuning with Adapters}

Fine-tuning pretrained foundational language models with adapter methodologies was popularized with the release of the ALPACA language model \cite{alpaca}. The objective of adapter methodologies, including the popular LoRA (low-rank adapter) method, is to avoid “catastrophic forgetting,” thereby allowing the language model to learn new information while retaining the prior info that it was pretrained upon \cite{Hu2021LoRALA}. In our setting, we are creating highly specialized language models against a new and challenging task, thus forgetting their prior information is likely acceptable in exchange for them performing well on our specialized task.

We applied the LoRA fine-tuning approach to the OPT-125M and OPT-1.3B pretrained foundational language models, along with our 100,000-example instruction fine-tuning data sets. In our studies, the LoRA parameter-efficient fine-tuning hyperparameters were $r=8$, $\alpha=32$, $lora\_dropout=0.05$, and $bias="none"$. 

Applying the LoRA fine-tuning adapters to all linear layers of the attention heads (1.37\% and 0.57\% of all tunable parameters, for OPT-125M and OPT-1.3B, respectively), resulted in poor performance of the LoRA fine-tuned language models relative to the performance of the instruction fine-tuned results (Table 3). Applying the LoRA fine-tuning adapters to only the attention $Q_{proj}$ and $V_{proj}$ tunable parameters for the OPT family of language models resulted in even lower performance (data not shown).

\begin{table*}[h!]
\begin{center}
\begin{small}
\begin{tabular}{lccccc}
\toprule
\specialcellcenter{Pretrained \\ Foundational \\ Language Model} & \specialcellcenter{Fine-Tuning \\ Method} & 
\specialcellcenter{\% Trainable \\ Model \\ Parameters} & \specialcellcenter{\% Exact \\ Matches} & \specialcellcenter{Mean \\ Normalized \\ Edit Distance} & \specialcellcenter{Mean \\ BLEU \\ Score} \\
\midrule
OPT-125M & LoRA	& 1.37\% & 0\% & 0.19 & 0.01 \\
OPT-125M & Standard & 100\% & 0\% & 0.50 & 0.20 \\
OPT-1.3B & LoRA	& 0.57\% & 0\% & 0.34 & 0.04 \\
OPT-1.3B & Standard & 100\% & 21\% & 0.75 & 0.63 \\
\bottomrule
\end{tabular}
\end{small}
\caption{Performance of standard instruction fine-tuned and LoRA instruction fine-tuned (adapters on all linear attention layers, $r = 8$, $\alpha = 32$) OPT language models with 100,000 instruction fine-tuning examples where learning rate = 2e-5, batch size = 4, and epochs = 3. Performance of the instruction fine-tuned language models were assessed with 1,000 test instances of SMILES strings, and the model’s ability to accurately generate the corresponding IUPAC chemical names relative to the ground truth.}
\end{center}
\end{table*}

\subsection{Importance of Instruction Data Formatting}

Next, we evaluated the formatting of the instruction fine-tuning data sets, and the downstream impact on our evaluation metrics. To explore this paradigm, the OPT-125M pretrained language model was fine-tuned on instruction data sets where the instructions were systematically inverted. The goal was to determine if providing the relevant information during language model fine-tuning, in any context, resulted in the desired outcome, or rather if specific formatting was required to achieve the desired outcome. 

Our data set formatting analysis utilized the same 1,000,000 example instruction fine-tuning dataset cohort employed throughout in this paper to maintain content consistency and control for this variable, while systematically inverting the instructions by 25\%, 50\%, 75\%, and 100\%. 

As an example, a standard format was:
\begin{itemize}
\item Instruction: “Translate the following SMILES string into an IUPAC name: \linebreak CN=C(C1=C(N=C(N1C)C2CCCCC2)C3=CC4=C(N3)C=CNC4=O)N,”, and
\item Output: “2-cyclohexyl-N',3-dimethyl-5-(4-oxo-1,5-dihydropyrrolo[3,2-c]pyridin-2-yl)imidazole-4-carboximidamide”.
\end{itemize}

\emph{Inversion} of the instruction format resulted in an example such as:
\begin{itemize}
\item Instruction: "Translate the following IUPAC name into a SMILES string: 2-cyclohexyl-N',3-dimethyl-5-(4-oxo-1,5-dihydropyrrolo[3,2-c]pyridin-2-yl)imidazole-4-carboximidamide,” and
\item Output: “CN=C(C1=C(N=C(N1C)C2CCCCC2)C3=CC4=C(N3)C=CNC4=O)N”.
\end{itemize}

Our systematic inversion study demonstrated the importance of proper instruction fine-tuning data formatting. The baseline, with 0\% inversion (\emph{i.e.}, standard formatting), demonstrated the best performance, with progressive erosion of fine-tuned model performance on our task as we increased the \% inversion. A 100\% inversion of the data set led to total ablation of task performance, as measured by both mean BLEU score and percentage of exact matches (Figure 8). 

\begin{figure}[h!]
\begin{center}
\includegraphics[width=100mm]{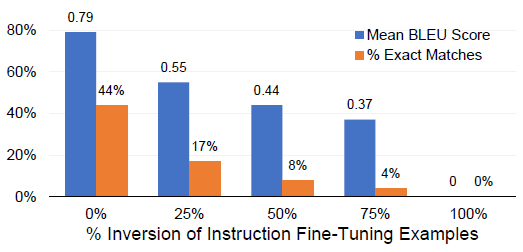}
\caption{Inversion of instruction fine-tuning examples and impact on evaluation metrics. Mean BLEU score (blue) and \% exact matches (orange) for the systematic inversion of the 1,000,000 instruction fine-tuning examples for the OPT-125M pretrained foundational language models, instruction fine-tuned with learning rate = 2e-5, batch size = 4, and epochs = 3. Performance of the instruction fine-tuned language models were assessed with 1,000 test instances of SMILES strings, in the standard (\emph{i.e.}, non-inverted) format, and the model’s ability to accurately generate the corresponding IUPAC chemical names relative to the ground truth.}
\label{pct_inversion_original_task}
\end{center}
\end{figure}

These results clearly demonstrated that proper formatting of the instruction fine-tuning data was paramount in achieving the desired outcome. Based on these outcomes, our recommendation is that the instruction fine-tuning data set should be formatted to reflect the intended utilization of the specialist fine-tuned language model.

\subsection{Inverse Task}

In addition to exploring the role of inverted fine-tuning data on our standard task, we also investigated the application of our instruction fine-tuning paradigm to the inverse task. Specifically, can a small pretrained foundational language models can be instruction fine-tuned to learn the task of converting IUPAC chemical names into the corresponding SMILES strings? 

We demonstrated that this inverse task is possible and with higher efficacy (Figure 9) than our standard task (\emph{see} Figure 5). The improved performance of our method on the inverse task is likely due to the constrained “vocabulary” of the inverse task outputs in comparison to our standard task outputs. For example, the IUPAC chemical names in our parent data set contained 55 unique characters whereas the SMILES strings contained 36 unique characters (35\% fewer characters than the IUPAC set).

\begin{figure}[h!]
\begin{center}
\includegraphics[width=100mm]{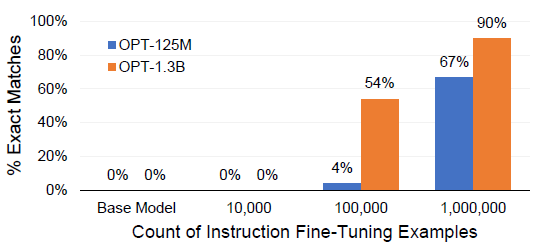}
\caption{Inverse task performance $w.r.t.$ \% exact matches for a range of instruction fine-tuning examples with the OPT-125M (blue) and OPT-1.3B (orange) pretrained foundational language models, instruction fine-tuned with learning rate = 2e-5, batch size = 4, and epochs = 3. Performance of the instruction fine-tuned language models were assessed with 1,000 test instances of IUPAC chemical names and the model’s ability to accurately generate the corresponding SMILES strings relative to the ground truth.}
\label{inverse_task}
\end{center}
\end{figure}

\section{Conclusion}

We have provided a framework for rigorous and systematic evaluation of our instruction fine-tuned language model outputs. We demonstrated that pretrained foundational language models, and their architectures, can serve as general learning frameworks for a novel task of which the base model is incapable of performing. 

Specifically, we demonstrated that 10,000-to-1,000,000 instruction examples provide optimal fine-tuning outcomes when paired with pretrained foundational language models of 125M-to-1.3B parameters. Successive language model instruction fine-tuning epochs can impart additional performance improvements and might serve as a surrogate strategy to improve fine-tuning performance with a smaller data set when a larger data set is not available. Notably, successive fine-tuning epochs are not infinitely additive, thus instruction fine-tuning for more than 20-30 epochs might provide diminishing returns.

Additionally, we demonstrated that very small language models from the TinyStories pretrained foundational language model family can be instruction fine-tuned to perform our task with measurable improvement over the baseline pretrained language models. Further, our approach can be generalized across many different open source software (OSS) pretrained foundational language models, although we noted performance differences that might be attributable to disparate foundational model architectures, attention strategies, and/or the information inherent within the pretrained foundational language models. 

Importantly, we explored the formatting of the instruction fine-tuning data set. We found the formatting of the instruction fine-tuning examples must match the fine-tuning objective to deliver optimal performance.

Although we achieved comparable, or superior, performance of our specialized task relative to other methods, as assessed by the mean BLEU score and \% exact matches, we do not advocate our process as a replacement for the state-of-the-art rules-based methods when converting SMILES strings into IUPAC chemical names or vice-versa. Conversely, our results serve as an example that it is possible to utilize a pretrained foundational language model in the same manner as a classical machine learning algorithm to achieve high-performance on a specialized task. In conclusion, we have demonstrated that pretrained generative language models can serve as general learning frameworks for sequence-based tasks.

\section{Acknowledgements}

The author would like to thank Anas Bricha and Neil Cameron for supporting this project. The author would also like to thank Guy Laporte for providing access to the computational infrastructure to conduct these studies. The author declares no financial interests nor conflicts. 

\newpage

\bibliography{bib-llms}
\bibliographystyle{icml2020}

\newpage

\appendix

\setcounter{table}{0}
\renewcommand{\thetable}{A\arabic{table}}

\section{Appendix}
\subsection{Computational Infrastructure and Code}

The results described in this article were carried out using a Dell Technologies PowerEdge XE9680 server with 8 x H100 NVIDIA\textsuperscript{\textregistered} HGX GPU cards with 80 GB VRAM each and NVLink\textsuperscript{TM} connectivity. There were 4 TB of CPU RAM, and 2 x Intel\textsuperscript{\textregistered} Xeon\textsuperscript{\textregistered} processors on the server. 

The server was configured with the Ubuntu v22.04 Linux operating system, Anaconda v23.1.0, NVIDIA\textsuperscript{\textregistered} CUDA v12.2, and NVIDIA\textsuperscript{\textregistered} drivers v535.129.03. Additional python dependencies included: \texttt{accelerate v0.25.0}, \texttt{bitsandbytes v0.41.3.post2}, \texttt{flash-attn v2.3.6}, \texttt{Levenshtein v0.23.0} (\emph{via} python-Levenshtein), \texttt{peft v0.7.1}, \texttt{scikit-learn v1.3.2}, \texttt{torch v2.1.2+cu121}, \texttt{transformers v4.36.2}, and \texttt{trl v0.7.4}.

The Stanford ALPACA language model code was git cloned directly from \url{https://github.com/tatsu-lab/stanford_alpaca} (accessed 30Dec2023). The \texttt{train.py} file in the GitHub repo, along with our corresponding instruction fine-tuning dataset, was used to instruction fine-tune the language models in our study. The language model fine-tuning code was executed via the command line interface (CLI).

Parameter-efficient fine-tuning (PEFT) of pretrained foundational language models with LoRA was conducted with the HuggingFace \texttt{transformers}, \texttt{peft}, and \texttt{trl} dependencies, as described at: \url{https://github.com/huggingface/peft} (accessed 30Dec2023) with \texttt{trl.SFTTrainer()}.

As an example, the following CLI command was used to instruction fine-tune a pretrained foundational language model on 8 GPUs:

\begin{lstlisting}
torchrun --nproc_per_node=8 [TRAINING_PY_FILE] \
    --model_name_or_path [HUGGINGFACE_MODEL_NAME] \
    --data_path [DATA_PATH_TO_FORMATTED_JSON_FILE] \
    --bf16 True \
    --output_dir [OUTPUT_DIRECTORY] \
    --overwrite_output_dir True \
    --num_train_epochs 3 \
    --per_device_train_batch_size 4 \
    --per_device_eval_batch_size 4 \
    --gradient_accumulation_steps 8 \
    --save_strategy "steps" \
    --save_total_limit 1 \
    --learning_rate 2e-5 \
    --weight_decay 0. \
    --warmup_ratio 0.03 \
    --lr_scheduler_type "cosine" \
    --seed 42 \
    --logging_steps 1 \
    --logging_dir ./logs \
    --tf32 True
\end{lstlisting}

\newpage

\subsection{Data Set Creation}

Our parent data set was composed of approximately 30 million instances of SMILES strings and their corresponding IUPAC chemical names from the free and publicly accessible United States National Institutes of Health (NIH) PubChem database. On 08Nov2023 116,109,741 compounds were available in PubChem \footnote{\url{https://ftp.ncbi.nlm.nih.gov/pubchem/Compound/Extras/} (accessed 08Nov2023).}. In applying the filtering criteria described below, the data set was reduced to 30,234,960 organic compounds (26\% of the original set). 

The following criteria were applied to create the filtered $\sim$30M organic compound data set: 
\begin{itemize}
    \item SMILES strings and IUPAC chemical names present;
    \item Deduplicated both SMILES and IUPAC fields; 
    \item SMILES contained only H, C, N, O, S, F, Cl, and/or Br elements;
    \item No isotopes, adducts, nor salts; 
    \item Removed unwanted functional groups such as nitro, nitroso, isonitrile, isocyanide, anhydride, epoxide, aziridine, azide, alkyl halide, acyl halide, sulfonyl halide, and/or 1,2-dicarbonyl;
    \item Molecular weight (MW) of 150 < MW < 550 daltons; 
    \item Fraction $sp^3$ (fsp3) $\ge 0.3$;
    \item Number of phenyl rings (nPh) $\le 2$;
    \item Number of aromatic rings (nAr) $\le 4$;
    \item Number of rings (nRings) $\ge 1$;
    \item Formal charge of zero;
    \item Number of rotatable bonds (nRot) $\ge 3$;
    \item Total polar surface area (TPSA) of 25 < TPSA < 150 \r{A}\textsuperscript{2};
    \item Calculated log P (cLogP) value for octanol/water partition of -2 < cLogP < 4.5; 
    \item Hydrogen bond donors (HBD) $\ge 4$. 
\end{itemize}

The above filtering criteria were applied with the \texttt{RDKit} python library for cheminformatics\footnote{\url{https://www.rdkit.org/}(accessed 30Dec2023)}. The selection criteria were consistent with best practices in filtering organic compound data sets for medicinal chemistry campaigns\footnote{Bunally, S. B.; Luscombe, C. N.; Young, R. J. "Using Physicochemical Measurements to Influence Better Compound Design." \emph{SLAS Discov.} \textbf{2019}, \emph{24}(\textbf{8}), 791-801.}. 

The pretrained foundational language models and the fine-tuned language models were evaluated against the test data. The same batch of test data was used for the evaluation of all the models in our study for consistency and comparison of outcomes.

\newpage

\subsection{Language Model Text Generation Configuration}

The same language model fine-tuning and generation configurations were utilized throughout our studies, and only single-parameter changes were permitted, as annotated in the tables, when comparing methods. Language model text generation was conducted via the HuggingFace \texttt{transformers} library. Transformers \texttt{GenerationConfig()} was set to the default parameters, along with:
\begin{itemize}
    \item \texttt{num\_beams = 2}, 
    \item \texttt{repetition\_penalty = 1.3}, 
    \item \texttt{do\_sample = False} (for consistent output generation), 
    \item \texttt{early\_stopping = True}, 
    \item \texttt{max\_time = 10}, and 
    \item \texttt{length\_penalty = 0.4}.
\end{itemize}

The text generation prompt and the general prompt used in the language model fine-tuning process were identical: 
\\
"\texttt{Below is an instruction that describes a task. Write a response that appropriately completes the request. \#\#\# Instruction: \{instruction\} \#\#\# Response: \{output\}}". 

Both the OPT and Bloom families of models required truncation of the output to the text following the "\texttt{\#\#\# Response:}" string. Similarly, the LLaMA-2 and Mistral families of models required truncation of the output to the text following the "\texttt{\#\#\# Instruction:}" string. Alternatively, the GPT-2, GPT-neo, and TinyStories families of models required truncation of the output to the text following the "\texttt{<|endoftext|>}" string.

\begin{table*}[h!]
\begin{center}
\begin{small}
\begin{tabular}{cccc}
\toprule
\specialcellcenter{Beam Size \\ (num\_beams)} & \specialcellcenter{\% Exact \\ Matches} & \specialcellcenter{Mean \\ Normalized \\ Edit Distance} & \specialcellcenter{Mean \\ BLEU \\ Score} \\
\midrule
1* & 27\% &  0.78  & 0.67 \\
2 & 44\% & 0.85 & 0.79 \\
3 & 44\% & 0.85 & 0.79 \\
4 & 44\% & 0.85 & 0.79 \\
5 & 44\% & 0.85 & 0.79 \\
6 & 44\% & 0.85 & 0.79 \\
\bottomrule
\end{tabular}
\end{small}
\caption{Variation of beam size and influence on text generation. Performance of an OPT-125M language model instruction fine-tuned on 1,000,000 examples with learning rate = 2e-5, batch size = 4, and epochs = 3. Performance of the instruction fine-tuned language models were assessed with 1,000 test instances of SMILES strings, and the model’s ability to accurately generate the corresponding IUPAC chemical names relative to the ground truth. *NOTE: when \texttt{num\_beams = 1}, \texttt{early\_stopping} and \texttt{length\_penalty} must also be disabled in \texttt{GenerationConfig()}.}
\end{center}
\end{table*}

A study to assess the impact of the \texttt{num\_beams} parameter on the \texttt{GenerationConfig()} command was conducted with the OPT-125M pretrained foundational language model that was fine-tuned on 1,000,000 instruction examples. The outcome of this study demonstrated that \texttt{num\_beams $\ge 2$} resulted in nearly identical outcomes, whereas \texttt{num\_beams = 1} revealed a much lower performance on all our evaluation metrics (Table A1). Based on this analysis, \texttt{num\_beams = 2} was selected for all text generation in our studies.

\newpage

\begin{table*}[t]
\begin{center}
\begin{small}
\begin{tabular}{cccc}
\toprule
\specialcellcenter{max\_time \\ (sec)} & 
\specialcellcenter{\% Exact \\ Matches} & \specialcellcenter{Mean \\ Normalized \\ Edit Distance} & \specialcellcenter{Mean \\ BLEU \\ Score} \\
\midrule
1 & 44\% & 0.85 & 0.79 \\
5 & 44\% & 0.85 & 0.79 \\
10 & 44\% & 0.85 & 0.79 \\
15 & 44\% & 0.85 & 0.79 \\
20 & 44\% & 0.85 & 0.79 \\
25 & 44\% & 0.85 & 0.79 \\
30 & 44\% & 0.85 & 0.79 \\
\bottomrule
\end{tabular}
\end{small}
\caption{Variation of \texttt{max\_time} and impact on text generation. Performance of an OPT-125M language model instruction fine-tuned on 1,000,000 examples with learning rate = 2e-5, batch size = 4, and epochs = 3. Performance of the instruction fine-tuned language models were assessed with 1,000 test instances of SMILES strings, and the model’s ability to accurately generate the corresponding IUPAC chemical names relative to the ground truth.}
\end{center}
\end{table*}

Further, \texttt{max\_time} was explored to assess the influence of this parameter on the \texttt{GenerationConfig()} command. The analysis was conducted with the OPT-125M pretrained foundational language model that was fine-tuned on 1,000,000 instruction examples. The outcome of this study demonstrated that \texttt{max\_time $\ge 1$} resulted in identical outcomes (Table A2). Based on this analysis, \texttt{max\_time = 10} was selected for all text generation in our studies to allow for any potential increase in text generation time required at inference with language models of >125M parameters.

\newpage

\subsection{Variability in Language Model Text Generation}

A single instruction fine-tuned language model was evaluated against a single test data set of 1,000 examples, in triplicate, to determine text generation variability and its impact on downstream evaluation metric variability. There was no detectable variability in the evaluation metrics (Table A3).

\begin{table*}[h!]
\begin{center}
\begin{small}
\begin{tabular}{cccc}
\toprule
\specialcellcenter{Inference} & 
\specialcellcenter{\% Exact \\ Matches} & \specialcellcenter{Mean \\ Normalized \\ Edit Distance} & \specialcellcenter{Mean \\ BLEU \\ Score} \\
\midrule
1 & 44\% & 0.85 & 0.79 \\
2 & 44\% & 0.85 & 0.79 \\
3 & 44\% & 0.85 & 0.79 \\
\bottomrule
\end{tabular}
\end{small}
\caption{Multiple inferences of the same instruction fine-tuned model and impact on text generation metrics. An OPT-125M language model instruction fine-tuned on 1,000,000 examples with learning rate = 2e-5, batch size = 4, and epochs = 3. Performance of the instruction fine-tuned language models were assessed with 1,000 test instances of SMILES strings, and the model’s ability to accurately generate the corresponding IUPAC chemical names relative to the ground truth.}
\end{center}
\end{table*}

Further, a single pretrained foundational language model was fine-tuned, in triplicate, using the same domain-specific training data set to determine the variability in model fine-tuning, and the downstream influence on text generation and metric evaluation variability. Again, there was no detectable variability in the evaluation metrics (Table A4). 

\begin{table*}[h!]
\begin{center}
\begin{small}
\begin{tabular}{cccc}
\toprule
\specialcellcenter{Language Model \\ Fine-Tuning Run} & 
\specialcellcenter{\% Exact \\ Matches} & \specialcellcenter{Mean \\ Normalized \\ Edit Distance} & \specialcellcenter{Mean \\ BLEU \\ Score} \\
\midrule
1 & 44\% & 0.85 & 0.79 \\
2 & 44\% & 0.85 & 0.79 \\
3 & 44\% & 0.85 & 0.79 \\
\bottomrule
\end{tabular}
\end{small}
\caption{Multiple fine-tuning runs to create the same instruction fine-tuned language model and assess the impact on text generation metrics. An OPT-125M language model instruction fine-tuned on 1,000,000 examples with learning rate = 2e-5, batch size = 4, and epochs = 3. Three separate fine-tuning runs were performed with these conditions. Performance of the instruction fine-tuned language models were assessed with 1,000 test instances of SMILES strings, and the model’s ability to accurately generate the corresponding IUPAC chemical names relative to the ground truth.}
\end{center}
\end{table*}

\newpage

Finally, a single pretrained foundational language model was fine-tuned using the same domain-specific training data set, using three discrete random seed values for the language model fine-tuning process. Our goal was to determine the variability in model fine-tuning with different random seeds, and the downstream impact on text generation and metric evaluation variability. In this setting there was no detectable variability in the normalized edit distance metrics and $\le 2\%$ variability in the percent exact match and mean BLEU score metrics (Table A5)

\begin{table*}[h!]
\begin{center}
\begin{small}
\begin{tabular}{cccc}
\toprule
\specialcellcenter{Language Model \\ Fine-Tuning Random \\ Seed Value} & 
\specialcellcenter{\% Exact \\ Matches} & \specialcellcenter{Mean \\ Normalized \\ Edit Distance} & \specialcellcenter{Mean \\ BLEU \\ Score} \\
\midrule
41 & 42\% & 0.85 & 0.78 \\
42 & 44\% & 0.85 & 0.79 \\
43 & 44\% & 0.85 & 0.79 \\
\bottomrule
\end{tabular}
\end{small}
\caption{Multiple runs to create the same instruction fine-tuned language model, with a different random seed value for each run, and assess the impact on text generation metrics. An OPT-125M language model instruction fine-tuned on 1,000,000 examples with learning rate = 2e-5, batch size = 4, and epochs = 3. Three separate fine-tuning runs were performed with these conditions, varying only the random seed value for the language model instruction fine-tuning process. Performance of the instruction fine-tuned language models were assessed with 1,000 test instances of SMILES strings, and the model’s ability to accurately generate the corresponding IUPAC chemical names relative to the ground truth.}
\end{center}
\end{table*}

\newpage

\subsection{Pretrained Language Model Baseline Performance}

All pretrained foundational language models explored in our analysis, along with some popular models that were not utilized in our study (\emph{e.g.}, LLaMA-2, Mistral, and Phi), were unable to perform our specialized task with any reasonable level of proficiency. A summary of the language model baseline performance is shown below with respect to our NLP metrics (Table A6).

\begin{table*}[h!]
\begin{center}
\begin{small}
\begin{tabular}{lcccc}
\toprule
\specialcellcenter{Pretrained Foundational \\ Language Model} & \specialcellcenter{Language Model \\ Parameter Count} & \specialcellcenter{\% Exact \\ Matches} & \specialcellcenter{Mean Normalized \\ Edit Distance} & \specialcellcenter{Mean BLEU \\ Score} \\
\midrule
roneneldan/TinyStories-1M & 1M & 0\% & 0.10 & 0 \\
roneneldan/TinyStories-3M & 3M & 0\% & 0.09 & 0 \\
roneneldan/TinyStories-8M & 8M & 0\% & 0.08 & 0 \\
roneneldan/TinyStories-1Layer-21M & 21M & 0\% & 0.08 & 0 \\
roneneldan/TinyStories-28M & 28M & 0\% & 0.02 & 0 \\
roneneldan/TinyStories-33M & 33M & 0\% & 0.08 & 0 \\
roneneldan/TinyStories-2Layers-33M & 33M & 0\% & 0.09 & 0 \\
EleutherAI/gpt-neo-125m & 125M & 0\% & 0.08 & 0 \\
gpt2 & 124M & 0\% & 0.06 & 0 \\
facebook/opt-125m & 125M & 0\% & 0.09 & 0 \\
facebook/opt-350m & 350M & 0\% & 0.07 & 0 \\
facebook/opt-1.3b & 1.3B & 0\% & 0.08 & 0 \\
facebook/opt-6.7b & 6.7B & 0\% & 0.03 & 0 \\
facebook/opt-13b & 13B & 0\% & 0.02 & 0 \\
facebook/opt-30b & 30B & 0\% & 0.01 &  0 \\
bigscience/bloom-1b1 & 1.1B & 0\% & 0.05 & 0 \\
bigcode/santacoder & 1.1B & 0\% & 0.04 & 0 \\
microsoft/phi-1\_5 & 1.5B & 0\% & 0.12 & 0 \\
microsoft/phi-2 & 2B & 0\% & 0.07 & 0 \\
meta-llama/Llama-2-7b-hf & 7B & 0\% & 0.00 & 0 \\
meta-llama/Llama-2-13b-hf & 13B & 0\% & 0.04 & 0 \\
meta-llama/Llama-2-70b-hf & 70B & 0\% & 0.24 & 0 \\
mistralai/Mistral-7B-Instruct-v0.2 & 7B & 0\% & 0.11 & 0 \\
\bottomrule
\end{tabular}
\end{small}
\caption{Baseline performance of pretrained foundational language models in the conversion of 1,000 test instances of SMILES strings into IUPAC chemical names, and comparison of the model output to the ground truth. The language models are described by their \texttt{HuggingFace.co} repo names (accessed 30Dec2023).}
\end{center}
\end{table*}

\newpage

\subsection{Examples of Variation IUPAC Names and Influence on Metrics}

Below are instances of SMILES strings and their corresponding IUPAC chemical names. There are also examples of IUPAC chemical name outputs provided by instruction fine-tuned language models, and the corresponding normalized edit distance and BLEU scores for those outputs relative to the ground truth data (Table A7).

\begin{table*}[h!]
\begin{center}
\begin{tiny}
\begin{tabular}{p{0.30\linewidth}p{0.25\linewidth}p{0.25\linewidth}cc}
\toprule
SMILES String & IUPAC Chemical Name & \specialcellcenter{IUPAC Chemical Name Provided \\ by Language Model} & \specialcellcenter{Normalized \\ Edit Distance} & \specialcellcenter{BLEU \\ Score} \\
\midrule
CCNCC1=CN(N=C1C(F)(F)F)C(C)C & N-[[1-propan-2-yl-3-(trifluoromethyl)pyrazol-4-yl]methyl]ethanamine & Write a response that appropriately completes the request. & 0.15 & 0 \\
CCC(CCCSC1=C(C=CC(=C1)F)F)(C\#N)NCC & 5-(2,5-difluorophenyl)sulfanyl-2-ethyl-2-(ethylamino)pentanenitrile & 3-methylphenylacetamide, 1-(2-methoxyethyl)-4-[5-(trifluoromethylsulfonyl)]imidazolecarboxylic acid amide & 0.23 & 0 \\
CN1CCOC(C1)CN2CCOC(C2)CN & 4-[(4-methylmorpholin-2-yl)methyl]morpholin-2-yl]methanamine & N-methylpyridinium chloride & 0.28 & 0 \\
CCC1=C(N(C(=O)CS1)C(C)C(=O)O)C2=CC=C(C=C2)C	& 2-[6-ethyl-5-(4-methylphenyl)-3-oxo-1,4-thiazin-4-yl]propanoic acid & 4-butyl-3-[(4-methylphenyl)methylsulfonyl]benzoic acid & 0.51 & 0.14 \\
CC1C(N(CCC1=O)CCC2=CC=CC3=C2N=CC=C3)C & 2,3-dimethyl-1-(2-quinolin-8-ylethyl)piperidin-4-one & 2-(4-methyl-1,2,4-triazol-5-yl)piperidin-2-yl & 0.52 & 0 \\
CC1=C(OC(=N1)C)CC(=O)NCCC2=CC=C(C=C2)Cl & N-[2-(4-chlorophenyl)ethyl]-2-(2,4-dimethyl-1,3-oxazol-5-yl)acetamide & 2-[(3-chlorophenyl)methyl]-N-(2-methylphenyl)acetamide & 0.65 & 0 \\
CC1CCCN(C1)C(=O)NC2CCN(CC2)CC3=CC=C(C=C3)Cl & N-[1-[(4-chlorophenyl)methyl]piperidin-4-yl]-3-methylpiperidine-1-carboxamide	& N-[(4-chlorophenyl)methyl]-2-(2-methylpropyl)piperidine-3-carboxamide & 0.68 & 0.35 \\
C1CN(CCC1C2=CSC=C2)S(=O)(=O)C3=CC=CC(=C3)C(F)(F)F & 4-thiophen-3-yl-1-[3-(trifluoromethyl)phenyl]sulfonylpiperidine & 2-methylsulfonyl-N-[1-(trifluoromethyl)phenyl]sulfonyl]piperidin-4-amine & 0.69 & 0.34 \\
COC1=CC=CC(=C1)[C@@H](C2CCN(CC2)CC3=CC=CC=C3)O & (R)-(1-benzylpiperidin-4-yl)-(3-methoxyphenyl)methanol & 2-[(3-benzylpiperidin-4-yl)methoxyphenyl)methanol & 0.83 & 0.27 \\
CCCOC1=CC=CC(=C1)C(CCO)C(CC)N & 4-amino-3-(3-propoxyphenyl)hexan-1-ol & 2-amino-3-(4-propoxyphenyl)propan-1-ol & 0.84 & 0.3 \\
CC1=CN=C(S1)CNS(=O)(=O)C2CCNCC2 & N-[(5-methyl-1,3-thiazol-2-yl)methyl]piperidine-4-sulfonamide & N-[(4-methyl-1,3-thiazol-5-yl)methyl]piperidine-1-sulfonamide & 0.95 & 0.52 \\
CC1=CN=C(S1)CNS(=O)(=O)C2CCNCC2 & N-[(5-methyl-1,3-thiazol-2-yl)methyl]piperidine-4-sulfonamide & N-[(4-methyl-1,3-thiazol-2-yl)methyl]piperidine-4-sulfonamide & 0.98 & 0.84 \\
C1CN(CCC1O)CCC2CNC2 & 1-[2-(azetidin-3-yl)ethyl]piperidin-4-ol & 1-[2-(azetidin-3-yl)ethyl]piperidin-4-ol & 1 & 1 \\
\bottomrule
\end{tabular}
\end{tiny}
\caption{SMILES strings and their corresponding IUPAC chemical names. There are also examples of IUPAC chemical name outputs provided by instruction fine-tuned language models, and the corresponding normalized edit distance and BLEU scores for those outputs relative to the ground truth data (\emph{i.e.}, "IUPAC Chemical Name" field in the table).}
\end{center}
\end{table*}

Normalized Levenshtein edit distance was used to compare the output of the language model against the ground truth data. Edit distance considers the number of deletions, substitutions, and transpositions required to transform one string into another \cite{Levenshtein1965BinaryCC}. Normalized edit distance is the edit distance between two strings, divided by the length of the longest string, thus bounded within $[0,1]$, where 1 is a perfect match \cite{Marzal1993ComputationON}. Normalized edit distance also allows for comparison of multiple results across a dataset when there is variation in string length.

We used the same prescriptive chunking and BLEU scoring $[0,1]$, where 1 is a perfect match, described in a previous publication \cite{Rajan2020STOUTST}. The chunked language model output was compared against the chunked ground truth via BLEU scoring.

\end{document}